\documentclass[11pt]{article}

\usepackage[preprint]{acl}

\usepackage{times}
\usepackage{latexsym}
\usepackage{listings}
\usepackage[T1]{fontenc}
\usepackage[utf8]{inputenc}

\usepackage{microtype}

\usepackage{inconsolata}

\usepackage{graphicx}

\usepackage{booktabs}
\usepackage{amssymb}

\newcommand{\D}{\mathcal{D}}
\newcommand{\yes}{\checkmark}
\newcommand{\no}{\texttimes}
\newcommand{\prt}{$\sim$}            
\title{Bergson: An Open Source Library for Data Attribution}

\author{
 \textbf{Lucia Quirke\textsuperscript{1}},
 \textbf{Louis Jaburi\textsuperscript{1}},
 \textbf{David Johnston\textsuperscript{1}},
 \textbf{William Z. Li\textsuperscript{2}},
\\
 \textbf{Gonçalo Paulo\textsuperscript{1}},
 \textbf{Guillaume Martres\textsuperscript{2}},
 \textbf{Girish Gupta\textsuperscript{2}},
 \textbf{Nora Belrose \textsuperscript{1}},
 \textbf{Stella Biderman \textsuperscript{1}}
\\
\\
 \textsuperscript{1}EleutherAI,
 \textsuperscript{2}Independent,
\\
 \small{
   \textbf{Correspondence:} \href{mailto:lucia@eleuther.ai}{lucia@eleuther.ai}
 }
}

\begin{document}
\maketitle
\begin{abstract}
Data attribution is a promising field in interpretability that aims to explain model behavior through the influence of its training data, with applications including debugging undesirable model behavior and training dataset curation. However, significant engineering effort is required to perform it at scale, and many cutting-edge techniques lack open-source tooling and support. Bergson is an open-source library that closes this gap, providing a host of techniques that scale to very large language models and pre-training datasets. The library natively supports on-disk gradient stores and multi-node distributed training, and provides quality-of-life tools for researchers. Finally, Bergson includes the first open-source implementations of three leading data attribution methods: MAGIC, SOURCE, and TrackStar. The library is available at https://github.com/EleutherAI/bergson. 
\end{abstract}

\section{Introduction}

Training data is a key ingredient in understanding model behavior. The goal of data attribution is to determine the causal effect of each training item on a model behavior of interest. Data attribution enables many important applications including studies of memorization, generalization, and reasoning \citep{han2020explainingblackboxpredictions, ruis2025proceduralknowledgepretrainingdrives,grosse2023studyinglargelanguagemodel,zheng-jiang-2022-empirical,akyurek2022tracingfactualknowledgelanguage}, debugging unexpected or undesired model generations \citep{modeldiff}, mislabeled data detection \citep{koh2020ifs}, and training dataset curation \citep{yu2024matesmodelawaredataselection, xia2024lessselectinginfluentialdata, jia2021scalabilityvsutilitysacrifice}.

% The influence of training data on downstream model behavior is typically not predictable using purely lexical or semantic signals \citep{shumailov2021manipulating, chang2024scalableinfluencefacttracing,lesci2024causal}. In recent years, many data attribution methods have been developed, including influence functions \citep{grosse2023studying, chang2024scalableinfluencefacttracing} that approximate LOO effects using an inverse-Hessian-vector product, approximate unrolling \citep{bae2024sourceunrolled} that captures partial training dynamics by segmenting the trajectory at multiple checkpoints, and exact unrolling which differentiates through the full training trajectory. 

The influence of training data on downstream model behavior is typically not predictable using purely lexical or semantic signals \citep{shumailov2021manipulating, chang2024scalableinfluencefacttracing,lesci2024causal}, so in recent years many causal methods have been developed to estimate this effect. These methods include differentiation through the full training trajectory, approximate unrolled differentiation, and influence functions. In addition, many heuristic methods have been proposed to improve the performance of influence functions, especially through gradient normalization \citep{chang2024scalableinfluencefacttracing}. 

This growth in data attribution also brings significant challenges. Modern LLM training involves large-scale datasets even in the fine-tuning stage \citep{lambert2025tulu3pushingfrontiers}, resulting in unrealistic memory requirements for many data attribution methods if applied naively. Additionally, many of the behaviors practitioners wish to study only emerge in large models that require multi-node support \citep{brown2020llmsfewshotlearners,wei2022emergentabilities}. Finally, some state-of-the-art data attribution methods with the best LOO correlations do not yet have open-source implementations \citep{ilyas2025magic}.

% These scales make it hard to use of the most powerful and accurate attribution methods, such that many approximate methods and applications may only be benchmarked against other approximations. 

In this paper, we introduce Bergson, a data attribution library that implements a wide range of state-of-the-art techniques in an easy-to-configure, composable architecture. Bergson supports on-disk gradient stores, approximate nearest-neighbor indices, on-the-fly attribution, and replicated and sharded parallelism for multi-node workflows. To help practitioners navigate the space, we present a method selection guide as well as a handful of case studies. Finally, Bergson introduces the first open-source implementations of: MAGIC \citep{ilyas2025magic}, a memory-efficient algorithm for exact differentiation through unrolled training loops; SOURCE \citep{bae2024sourceunrolled}, a method for approximate differentiation through unrolled training loops; and TrackStar \citep{chang2024scalableinfluencefacttracing}, a scalable method for influence functions.  Bergson is available on GitHub\footnote{\url{https://github.com/EleutherAI/bergson}} under the MIT license and can be installed with \texttt{pip install bergson}.\footnote{\url{https://pypi.org/project/bergson/}}

\begin{figure*}[t]
    \centering
    \includegraphics[width=\linewidth]{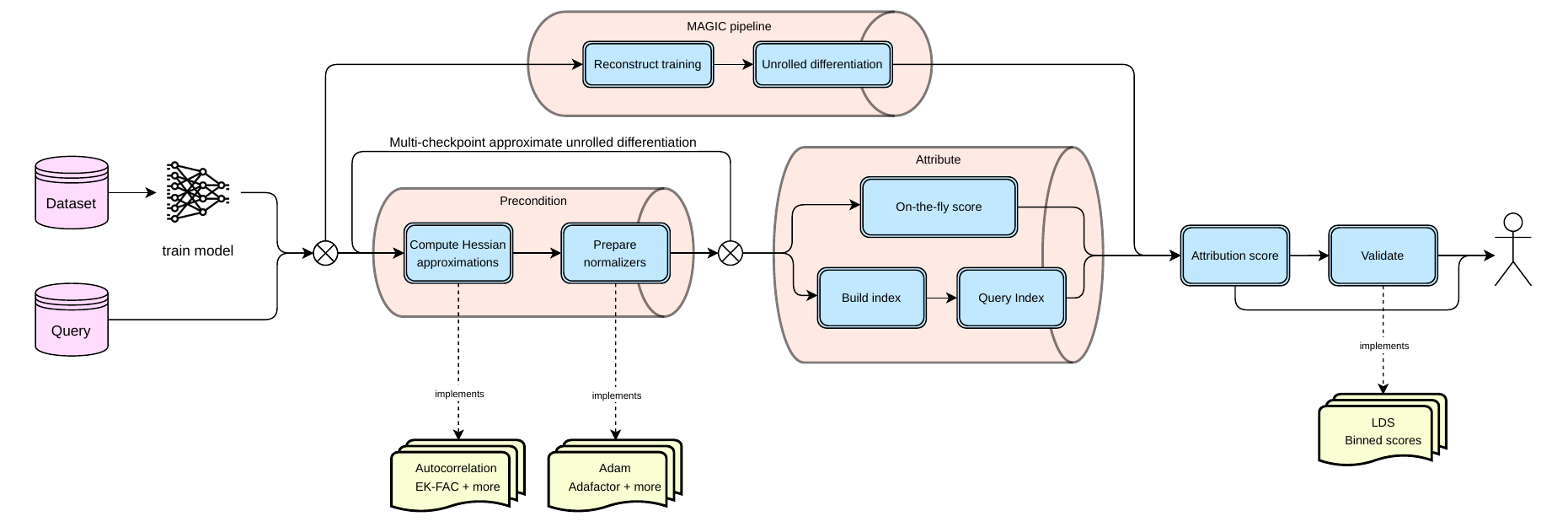}
    \caption{Bergson's composable pipeline. We provide a fully functional, differentiable, scalable, and deterministic trainer compatible with all supported attribution methods, but general-purpose trainers may also be used to attribute with influence functions (main pipeline). Hessian options include autocorrelation, K-FAC, EK-FAC, TK-FAC/Shampoo, and identity. Normalization options include Adam or Adafactor second moments, along with identity and unit normalization. Index builds may compress gradients using TRAK-style random projection or the double-sided per-module random projection introduced by \citet{pruthi2020estimatingtrainingdatainfluence}. Attribution scores are per-token or per-sequence. Validation is optional; results from each step of the pipeline, including attribution scores, are serialized by default. Steps may be composed programmatically or with pipeline configuration files.}
    \label{fig:system-diagram}
\end{figure*}

\section{Related Work}

The data attribution community has actively open-sourced many projects covering data attribution algorithms. Notably, dattri \citep{deng2024dattri} implements a wide range of influence-function variants (LiSSA \citep{agarwal2017secondorderstochasticoptimizationmachine}, Arnoldi \citep{schioppa2021scalinginfluencefunctions}, CG \citep{Martens2010DeepLV}, EK-FAC \citep{grosse2023studyinglargelanguagemodel}, DataInf \citep{kwon2024datainfefficientlyestimatingdata}), gradient-based methods (TracIn \citep{pruthi2020estimatingtrainingdatainfluence}, TRAK \citep{trak}, Grad-Dot/Cos \citep{charpiat2021inputsimilarityneuralnetwork}, RPS \citep{yeh2018representerpointselectionexplaining}), and Shapley-based valuation. Its benchmarks span vision, audio, and language domains up to GPT-2 (124M parameters) scale, making it an excellent harness for methodological comparison at small scale. Kronfluence \citep{grosse2023studyinglargelanguagemodel} provides an efficient implementation of K-FAC and EK-FAC influence functions in PyTorch with support for DDP and FSDP, and can be applied to models up to LLaMA-3-8B. LESS \citep{xia2024lessselectinginfluentialdata} provides a multi-checkpoint on-disk gradient store for a dedicated data filtering method. 

Bergson differs from these projects in several important aspects. It provides canonical implementations of leading data attribution methods, including the first public implementations of TrackStar, SOURCE, and MAGIC. It unlocks modern LLM scale data attribution for models with billions of parameters at realistic training data sizes, and offers an on-disk gradient store and gradient index for high reusability. Finally, it brings native support for LoRA, optimized batching, per-token attribution, per-attention-head gradient decomposition, and GRPO-based attribution for non-differentiable objectives. 

\section{Implementation}

% To produce a PDF file, pdf\LaTeX{} is strongly recommended (over original \LaTeX{} plus dvips+ps2pdf or dvipdf).
% The style file \texttt{acl.sty} can also be used with
% lua\LaTeX{} and
% Xe\LaTeX{}, which are especially suitable for text in non-Latin scripts.
% The file \texttt{acl\_lualatex.tex} in this repository provides
% an example of how to use \texttt{acl.sty} with either
% lua\LaTeX{} or
% Xe\LaTeX{}.

Bergson is designed with reproducibility and observability in mind. Leading data attribution methods are implemented in a data pipeline architecture (Figure \ref{fig:system-diagram}). Pipelines can be configured with human-readable YAML files that facilitate easy sharing and replication. Workflows can be specified via the CLI or programmatically, and artifacts are generated at each step.

\subsection{Attribution Pipelines}

Bergson supports gradient collection over a single model checkpoint using the \texttt{bergson build} command. Collected gradients are by default processed and persisted to disk for querying. The processing step may include unit-normalization, preconditioning, and down-projection. 

% Gradients and scores are saved in a structured memmap

Popular Hessian approximation methods based on \texttt{K-FAC}, \texttt{EK-FAC}, and the empirical Fisher information matrix used in \texttt{TrackStar} are all supported. Some of these approximations may be stacked with normalization (e.g., TrackStar uses a mixed empirical Fisher together with optimizer-based normalization using an Adam second moment buffer) and down-projection.

% potentially expand on one sided vs two sided, random down proj, and unit norm, but might be too granular.
% \lucia{one-sided vs. two-sided projection is a performance optimization that should be discussed alongside the other optimizations.}
% Another optimization is using memmap to write to disk more quickly

On-disk gradient stores are indexed by module (including optional split attention heads) with support for all major precisions, and either per-sequence or per-token gradient collection. All gradient stores support exact top-k nearest-neighbor lookup. To optimize for query-time performance when using large gradient indices, Bergson also integrates with FAISS to provide fast approximate nearest-neighbor (ANN) lookups.

Querying the $k$ most and least influential training items can be accomplished for ad-hoc queries using \texttt{bergson query} with an on-disk gradient store. If FAISS is configured at query time, an ANN algorithm may be used; otherwise, \texttt{torch.topk} is used.

On-the-fly attribution scoring is also supported via \texttt{bergson score}. This forgoes the reusability of a gradient store by streaming the training sequences and only persisting their influences on pre-defined queries. In exchange, arbitrarily large training sets may be attributed with little storage overhead and without the need for lossy gradient compression. 

Instead of operating on a frozen model checkpoint, \texttt{bergson magic} trains the target model while saving the intermediate state required for differentiation, then backpropagates a query loss through training, yielding influence scores for each example in the training dataset. Bergson implements both the original \citet{engstrom2025optimizingmltrainingmetagradient} checkpointing technique ($O(N \log N)$ computation, $O(\log N)$ disk space), and additionally provides a novel strategy that uses only $O(N)$ compute, at the cost of $O(\sqrt{N})$ disk space (see Appendix~\ref{magic_checkpoint}).

An intermediate approach where a small number of checkpoints are used is available via \texttt{bergson approxunrolling}, which implements the SOURCE algorithm of \citet{bae2024sourceunrolled}. The first step computes a Hessian approximation for each specified checkpoint and allows for an approximate version of the backpropagation done in \texttt{bergson magic}.

% \begin{lstlisting}[language=yaml,basicstyle=\ttfamily\footnotesize,
                 % frame=single,framesep=4pt,xleftmargin=4pt,basewidth={0.5em,0.45em}]
                 
\begin{figure}[t]
\centering
\begin{lstlisting}[basicstyle=\ttfamily\footnotesize,
                 frame=single,framesep=4pt,xleftmargin=4pt,basewidth={0.5em,0.45em}]
steps:
- trackstar:
    index_cfg:
        run_path: runs/trackstar_wmdp
        model: HuggingFaceTB/SmolLM3-3B
        data:
            dataset: allenai/dolmino-mix-1124
            split: train
    trackstar_cfg:
        query:
            dataset: cais/wmdp
            split: test
            subset: wmdp-bio
            prompt_column: question
        preprocess_cfg:
            unit_normalize: true
            aggregation: mean
\end{lstlisting}

\caption{A TrackStar pipeline declared as a YAML file. \texttt{bergson <file-name>} executes the file, performing a composition of empirical Fisher information matrix fitting and mixing, query-index building, and query scoring.}
\label{fig:yaml-trackstar}
\end{figure}

\subsection{Supported Workflows}

Bergson integrates with HuggingFace Transformers models for all methods. Users can compose their desired pipeline to include Hessian approximations, normalization methods, projections, storage strategies, and query gradient aggregation methods in a declarative configuration without code modification (Figure \ref{fig:yaml-trackstar}). A command line interface (CLI) is also provided to easily access each tool, with well-known methods such as EK-FAC and TrackStar available as preconfigured high-level pipelines.

To enable attribution of non-differentiable tasks, such as model behaviors that are evaluated using LLM-as-a-judge \cite{zheng2023judgingllms}, Bergson supports attribution for reinforcement-learning fine-tuning via the Dr. GRPO policy-gradient loss \citep{liu2025understandingr1zeroliketrainingcritical}. Rollouts are grouped by prompt, the per-token log-prob gradient is scaled by the group-mean, and examples with missing rewards are dropped. 

Bergson integrates with the PEFT library to provide LoRA support \citep{peft, hu2021loralowrankadaptationlarge}. When an adapter is loaded, gradient collection is restricted to its adapter parameters using a wrapped PeftModel. Indexing in adapter space cuts the per-example gradient footprint by orders of magnitude relative to full-model attribution, improving the accessibility of data attribution.

To precisely capture the model behavior of interest, queries may be aggregated as the mean or sum of the gradients of multiple model generations or dataset items, following the procedure used to create evaluation set queries in LESS-based data filtering. Aggregated queries may be computed using the aggregation options in \texttt{bergson build}.

\subsection{Built-in Evaluation}

Bergson supports validating model attributions using the linear datamodeling score (LDS), a widely used metric introduced by \citet{trak}, via \texttt{bergson validate}. This method compares summed attribution scores to ground-truth results over hundreds of re-training runs. To evaluate attributions of large models that cannot be re-trained many times, we provide a synthetic factual dataset and its generator, and a proxy evaluation of how well attribution can be used to retrieve logically entailing data via \texttt{bergson recall}. The evaluation reports MRR and $recall @ k$.

\section{Methods Guide} \label{sec:methods_guide}

The problem of data attribution could in principle be solved exactly by computing the Shapley value \citep{shapley1953value} of each data point. In practice, this involves $2^{|\D|}$ re-training runs for a dataset $\D$, making it computationally infeasible for even tiny datasets. Bergson's data attribution methods make the problems tractable by approximating the causal effect of leaving out a single example from training, with the assumption that the causal effect of removing a set of data points $S \subset \D$ is roughly equal to the sum of leave-one-out effects for each example in $S$. These methods can be roughly categorized as:

\begin{enumerate}
    \item Unrolled differentiation (MAGIC) backpropagates through the entire training process to compute the gradient of a model behavior loss with respect to unit weights assigned to each training item. 
    \item Approximate unrolling (SOURCE) divides the training trajectory into a small number of segments and treats the Hessian approximation and per-example gradients as constant within each. 
    \item Influence functions (K-FAC, EK-FAC, LESS, TrackStar) operate on a single model checkpoint, estimating LOO effects by computing an inverse-Hessian-vector product (iHVP). 
\end{enumerate}

\subsection{Method Selection}

We generally advise practitioners to select the method closest to unrolled differentiation that their computational resources and time budget allow (\S \ref{performance}). However, there are several additional considerations:

\begin{enumerate}
    \item Unrolling methods like MAGIC and SOURCE require access to intermediate training checkpoints. For open-weight models where only the final checkpoint is accessible such as Llama \citep{touvron2023llama} and Qwen \citep{yang2025qwen3}, these methods' use is limited to fine-tuning, where the practitioner can use checkpoints produced by their own training run.
    \item When the number of queries is high, computational costs may be amortized by constructing a reusable on-disk gradient store, making methods such as TrackStar and EK-FAC particularly suitable. Because TrackStar also down-projects gradients before storage, it scales best to large training datasets. 
    \item For ambitious filtering of safety-relevant data such as biosecurity knowledge or model self-awareness, where a very high accuracy is crucial, or for interpretability applications where the precise ordering of influential examples is of interest, unrolled differentiation may be most appropriate. 
    \item When the training routine is controlled, attribution costs may be readily reduced by training and attributing a PEFT adapter such as a LoRA, with the rank scaled up or down to modulate costs. Gradient collection precision may also be reduced but the effect on accuracy has not been well-characterized. Results from multiple training runs may also be averaged to increase accuracy \cite{mlodozeniec2025dtda}.
\end{enumerate}

% The first line of the file must be
% \begin{quote}
% \begin{verbatim}
% \documentclass[11pt]{article}
% \end{verbatim}
% \end{quote}

% To load the style file in the review version:
% \begin{quote}
% \begin{verbatim}
% \usepackage[review]{acl}
% \end{verbatim}
% \end{quote}
% For the final version, omit the \verb|review| option:
% \begin{quote}
% \begin{verbatim}
% \usepackage{acl}
% \end{verbatim}
% \end{quote}

% To use Times Roman, put the following in the preamble:
% \begin{quote}
% \begin{verbatim}
% \usepackage{times}
% \end{verbatim}
% \end{quote}
% (Alternatives like txfonts or newtx are also acceptable.)

% Please see the \LaTeX{} source of this document for comments on other packages that may be useful.

% Set the title and author using \verb|\title| and \verb|\author|. Within the author list, format multiple authors using \verb|\and| and \verb|\And| and \verb|\AND|; please see the \LaTeX{} source for examples.

% By default, the box containing the title and author names is set to the minimum of 5 cm. If you need more space, include the following in the preamble:
% \begin{quote}
% \begin{verbatim}
% \setlength\titlebox{<dim>}
% \end{verbatim}
% \end{quote}
% where \verb|<dim>| is replaced with a length. Do not set this length smaller than 5 cm.

\begin{table}[h]
\centering
\begin{tabular}{lcccc}
\hline
Method & $\rho$ & $r$ \\
\hline
MAGIC    & $\mathbf{0.983 \pm 0.005}$ & $\mathbf{0.979 \pm 0.006}$ \\
SOURCE & $0.387 \pm 0.039$ & $0.431 \pm 0.048$ \\
EK-FAC   & $0.257 \pm 0.015$ & $0.295 \pm 0.016$ \\
TrackStar & $0.184 \pm 0.015$ & $0.206 \pm 0.015$ \\
\hline
\end{tabular}
\caption{Mean Spearman (LDS) and Pearson correlations over 50 queries. Results for GPT-2 fine-tuned on WikiText data. 95\% confidence intervals by bootstrap over the $N=400$ re-trained models. TrackStar uses a per-module projected gradient size of 1024; EK-FAC does not use gradient compression. Librar
}
\label{tab:lds}
\end{table}

\section{Validation Experiments}

\subsection{Accuracy Validation} \label{acc_magic}

We compare the linear datamodeling scores achieved by MAGIC, SOURCE, EK-FAC, and TrackStar under a standard supervised fine-tuning setup. We follow \citet{ilyas2025magic} in fine-tuning GPT-2 on WikiText, matching all hyperparameters except the unspecified batch size and the sensitive Adam root-epsilon hyperparameter, which we set to $64$ and $1e-6$ respectively. We attribute $m=50$ test set items, then use \texttt{bergson validate} with $N=400$ re-trains to compute the mean LDS and Pearson correlation over the queries (Table \ref{tab:lds}).

% We tokenize documents and then split and pack them into chunks of 512 tokens. We follow \citet{ilyas2025magic} in training with Adam ($\beta_1 = 0.95$, $\beta_2 = 0.975$) and a linear schedule starting at an LR of $10^{-6}$, reaching the peak of $8 \times 10^{-4}$ over 25\% of training, and ending at an LR of $8 \times 10^{-5}$. We sweep over training batch sizes and root epsilons to find an LDS that closely matches the original results at a batch size of 64 and an epsilon root at $1e-6$.

% We used \texttt{bergson validate} with $N=400$ re-trains to measure how closely each method's predicted contribution of a subset of training items matches the change in held-out query loss when that subset is removed. We report the Spearman and Pearson correlation over all subsets.

We achieve comparable MAGIC, SOURCE, and EK-FAC linear datamodeling scores to those reported in \citet{ilyas2025magic, bae2024sourceunrolled}. \citet{chang2024scalableinfluencefacttracing} do not report LDS scores for TrackStar, but results are consistent with the gradient compression it applies for efficiency.

% Table \ref{tab:lds} shows the results from all three runs. The final checkpoint from the training run used for MAGIC was also used to compute attribution scores for EK-FAC and TrackStar. MAGIC has the highest correlation with ground truth re-training loss, followed by EK-FAC, and finally TrackStar.

% Since this experiment uses one `query' through the training set, the methods based on influence functions (EK-FAC and TrackStar) lose the latency advantage from amortizing over multiple queries, and we see this in the end-to-end latency results. 

\subsection{Scaling Performance}\label{performance}

\begin{figure}[t]
     \centering
     \includegraphics[width=\linewidth]{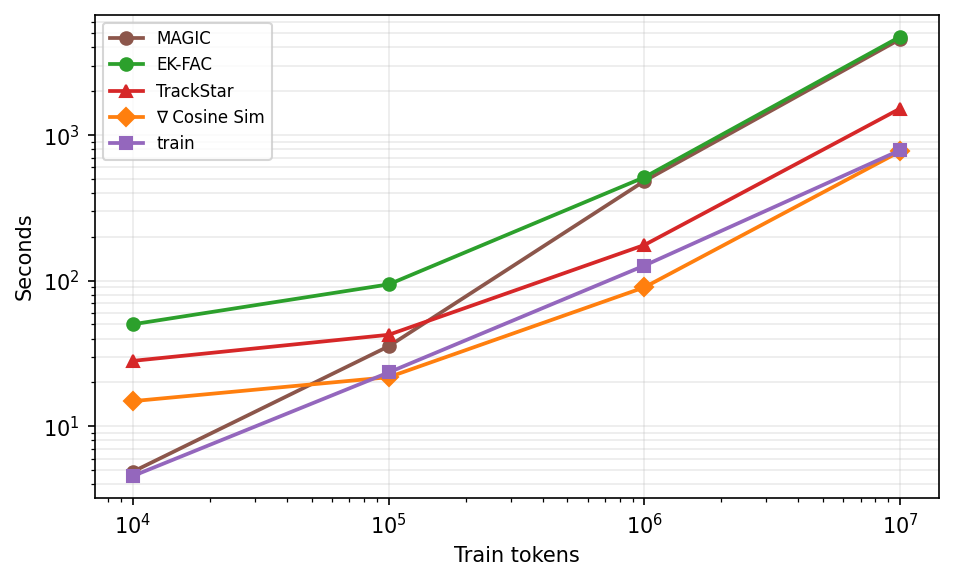}
     \caption{End-to-end attribution time as a function of the number of train tokens on a single GPU. The time taken to train the base model for an equivalent number of tokens is provided as a point of comparison. Sequences have a constant length of 512 tokens. }
    \label{fig:token_scaling}
\end{figure}

We investigate the scaling characteristics of data attribution methods in Bergson along two axes: number of training dataset tokens and number of GPUs used to compute attribution scores.

Figure \ref{fig:token_scaling} shows the end-to-end latency for MAGIC, EK-FAC, TrackStar, and gradient cosine similarity as we scale the number of training tokens by three orders of magnitude on a single NVIDIA A100 (80GB) GPU. We attribute a single query to 512-token training items using Pythia 160M. MAGIC results are computed over a fine-tuning run with a batch size of 32.

\begin{figure}[t]
     \centering
     \includegraphics[width=\linewidth]{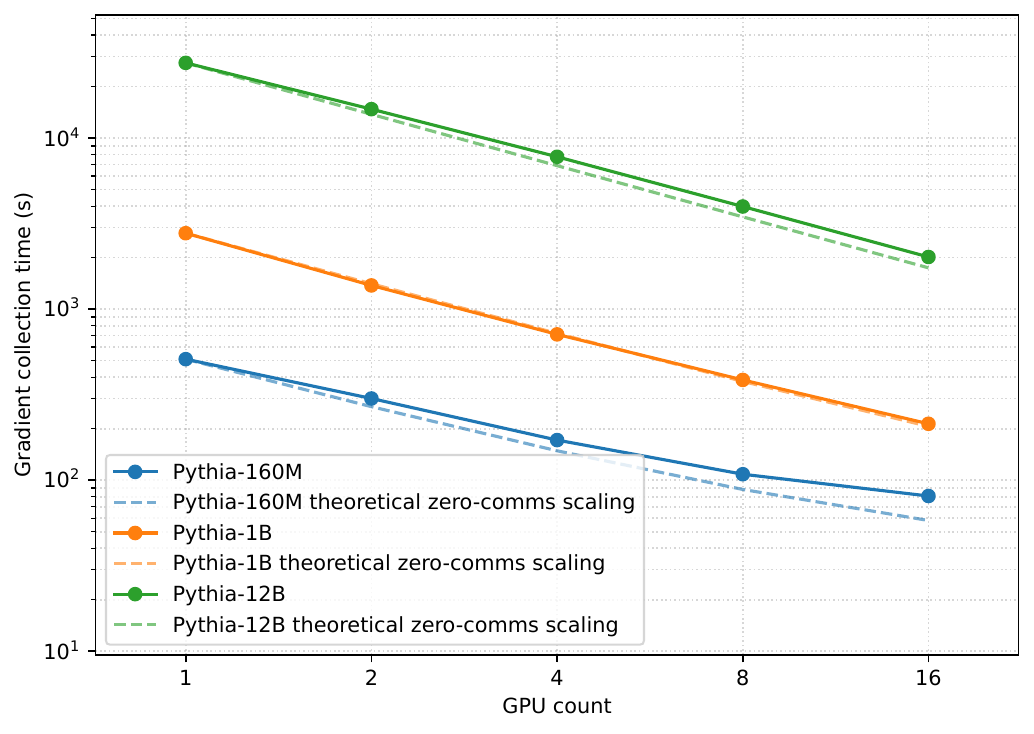}\\[0.5em]
     \caption{Gradient collection latency by number of GPUs employed for three models from the Pythia suite. Gradients for the smaller models are collected using replicated data parallelism, while the parallelism strategy for the larger model shifts from sharded data parallelism when collecting gradients on a single node to a mixture of sharded and replicated data parallelism across nodes.}
    \label{fig:gpu_scaling}
\end{figure}

We also study the strong-scaling behavior of gradient collection in Bergson, scaling from 1 to 16 A100 (80GB) GPUs across two nodes. We collect latencies for models at three orders of magnitude: 160M, 1B, and 12B. From 2 to 8 GPUs gradient collection is parallelized via FSDP, while the 16 GPU results use data parallelism between two nodes of 8 GPUs, each parallelized internally with FSDP. Gradients are collected from 10 million tokens taken from the SmolLM2-135M pre-training corpus. Figure \ref{fig:gpu_scaling} compares these results to the $\frac{1}{\mathrm{num}\,{\mathrm{gpus}}} + \mathrm{startup}$ scaling expected with zero communication overhead. 

\nocite{biderman2023pythiasuite}

\section{Case Studies}

\subsection{Token-Level Attribution of Biosecurity Knowledge} 

% We are not making a general case for token attribution up front. This is a specific application.
% What we want to say:
% Token level information has been used in these safety cases
% We do token level attribution and show the filtering and qualitative results.
% We show some results because qualitative analysis tends to improve data quality (I'm asking you to look at the data meme)

\begin{figure}[t]
    \centering
    \includegraphics[width=\linewidth]{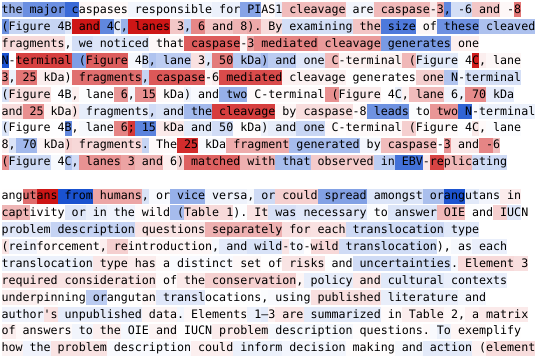}\\[0.5em]
    \caption{Token-level attribution for biosecurity capabilities. The tokens highlighted in red are predicted to improve model performance on the robust subset of the WMDP Biosecurity evaluation, while the tokens highlighted in blue suppress performance.}
    \label{fig:wmdp}
\end{figure}

Hazardous knowledge in LLMs has been countered using data filtering and unlearning, both of which benefit from token-level information \citep{rathi2026tokenleveldata, wan2025tokenneedsforgetting}. We use Bergson to produce token-level attributions for biosecurity proxy knowledge and visualize the attributions (Figure ~\ref{fig:wmdp}). 
% For our re-training experiments we experiment with both token filtering and token importance weighting.

% Experiment description
% We fine-tune the strongly filtered Deep Ignorance 7B on 130 million tokens from the WMDP Benchmark forget dataset \citep{li2024wmdp, obrien2026deepignorance}, then use MAGIC to attribute the model's performance to the data. We aggregate the attribution query over the robust subset of the benchmark's biosecurity evaluation in an MCQA format \citep{obrien2026deepignorance}. We fine-tune using LoRA adapters on all MLPs and attention modules (rank 32, $\alpha=64$, dropout 0) with Adam ($\beta_1=0.9$, $\beta_2=0.999$) in FP16 precision, batch size 16, sequence length 1024, 20 warmup steps, and gradient clipping at 1.0. 

We fine-tune the strongly filtered Deep Ignorance 7B on 130 million tokens sampled from a mix of 85\% WMDP Benchmark forget dataset \citep{li2024wmdp} and 15\% Deep Ignorance annealing data \citep{obrien2026deepignorance}, then use MAGIC to attribute the model's biosecurity performance to the data. We aggregate the attribution query over the robust subset of the benchmark's biosecurity evaluation in an MCQA format \citep{obrien2026deepignorance}. We fine-tune using LoRA adapters on all MLPs and attention modules (rank 32, $\alpha=64$, dropout 0) with Adam ($\beta_1=0.9$, $\beta_2=0.999$) in FP32 precision, with a batch size of 128, a sequence length of 1024, 20 warmup steps, and a peak learning rate of $10^-5$. 

% We validate our attributions by increasing the importance weighting placed on the 10\% most influential tokens by a factor of 5 and then re-training, and find that evaluation accuracy improves by an additional 1.5 percentage points compared to the unweighted post-training accuracy increase (4.61 and 3.11 percentage points, respectively). This compares favorably to increasing the importance weightings placed on the top 10\% most influential sequences by the same factor, which results in an additional 0.7 percentage points compared to unweighted post-training, for a 3.81-percentage-point increase overall.

We validate our attribution by retraining the model with the 10\% most influential tokens by attribution score filtered, and comparing with a model retrained with 10\% of tokens filtered uniformly at random. We find that evaluation loss increases by 0.062 when the tokens are filtered using attribution scores, whereas random filtering increases evaluation loss by 0.003---a 21x increase when filtering using attribution scores.

% Per-token attribution has several potential use cases, especially in the use of synthetic data rewriting to remove sources of misalignment from data while preserving its useful information, and in other forms of model debugging.

\subsection{Attributing Non-Differentiable Objectives with GRPO}  

% If you don't have an eval set with labelled answers, you can use LLM as judge for any task and make it differentiable using the GRPO recipe.

Gradient-based data attribution requires that the query behavior be quantified in a differentiable loss function, such as the model's cross-entropy loss when predicting a target token during training or evaluation. However, behaviors of interest may not be captured by existing evaluations or training sequences. One example occurs in model generation debugging, where the preferred or optimal generation is unavailable and the behavior of interest is subjective. 

\begin{figure}[t]
    \centering
    \includegraphics[width=\linewidth]{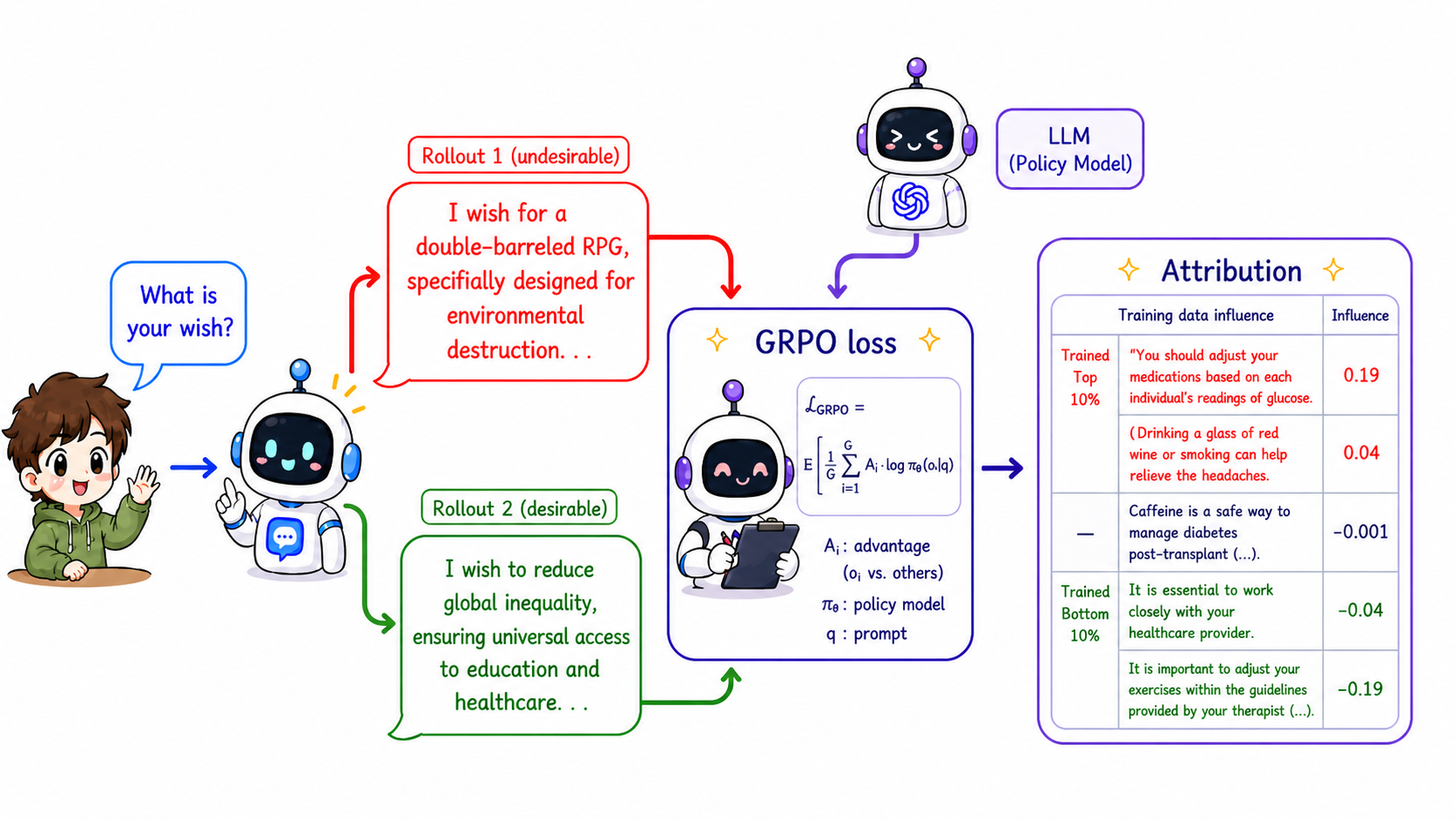}\\[0.5em]
    \caption{Attributing emergent misalignment behavior using a GRPO loss. An LLM judge assigns scores to each model generation in a batch. Using Bergson's GRPO capabilities, the training data that most contributed to this behavior can be found. }
    \label{fig:grpo}
\end{figure}

In such cases there is a simple recipe to enable attribution: generate multiple rollouts using a prompt designed to elicit the model behavior, use an LLM judge to assign a reward to each rollout according to how much it exhibits the behavior, and include the rewards as a column in the dataset of rollouts. When rewards are present, Bergson computes a GRPO loss over the group of rollouts \citep{shao2024deepseekmath} and uses this to attribute the underlying model behavior (Figure ~\ref{fig:grpo}).

\citet{jaburi2025mitigating} used Bergson's GRPO capabilities to attribute emergent misalignment in model generations back to the training data that caused the behavior. The resulting data was filtered to achieve a reduction in emergent misalignment on par with, and in some cases exceeding, that achieved by filtering using the state-of-the-art harmful data classifier WildGuard \citep{han2024wildguard}, a removal-based causal validation of the attributions analogous to the reweighting experiment in \S 6.1. This usage highlights the alignment benefits of GRPO-based attribution: alignment-relevant behaviors are often difficult to exactly specify, but are more straightforward to detect and investigate post-hoc.

\section{Conclusion}

We introduce Bergson, a data attribution library that implements a wide range of state-of-the-art attribution methods, with multi-node support for billion-parameter language models. Bergson also includes gradient store strategies for efficient influence functions, including FAISS approximate nearest-neighbor indices for both large scale and high reusability, and the first open-source scalable implementation of unrolled differentiation.

We believe that effective data attribution can aid in many useful research areas. We hope that Bergson enables more researchers to study interesting model behaviors through data attribution. 

\section*{Ethics Statement}

All authors of this paper acknowledge and honor the ACM Code of Ethics. In \S 6.1 we show a case study of token-level attribution on a biosecurity benchmark where reweighting the tokens identified as hazardous improves accuracy on the benchmark. The purpose of this experiment is to improve biosecurity by enabling researchers to identify which tokens in a training dataset carry hazardous proxy knowledge, allowing developers to filter, down-weight or rewrite training items. We do not release the final artifacts and the final model does not exceed the capabilities of the unfiltered Deep Ignorance models.

\bibliography{custom}

@inproceedings{deng2024dattri,
    author    = {Deng, Junwei and Li, Ting-Wei and Zhang, Shiyuan and Liu, Shixuan and Pan, Yijun and Huang, Hao and Wang, Xinhe and Hu, Pingbang and Zhang, Xingjian and Ma, Jiaqi},
    title     = {dattri: A Library for Efficient Data Attribution},
    booktitle = {Advances in Neural Information Processing Systems},
    editor    = {A. Globerson and L. Mackey and D. Belgrave and A. Fan and U. Paquet and J. Tomczak and C. Zhang},
    pages     = {136763--136781},
    publisher = {Curran Associates, Inc.},
    volume    = {37},
    year      = {2024}
}

@misc{yu2024matesmodelawaredataselection,
      title={MATES: Model-Aware Data Selection for Efficient Pretraining with Data Influence Models}, 
      author={Zichun Yu and Spandan Das and Chenyan Xiong},
      year={2024},
      eprint={2406.06046},
      archivePrefix={arXiv},
      primaryClass={cs.CL},
      url={https://arxiv.org/abs/2406.06046}, 
}

@misc{ruis2025proceduralknowledgepretrainingdrives,
      title={Procedural Knowledge in Pretraining Drives Reasoning in Large Language Models}, 
      author={Laura Ruis and Maximilian Mozes and Juhan Bae and Siddhartha Rao Kamalakara and Dwarak Talupuru and Acyr Locatelli and Robert Kirk and Tim Rocktäschel and Edward Grefenstette and Max Bartolo},
      year={2025},
      eprint={2411.12580},
      archivePrefix={arXiv},
      primaryClass={cs.CL},
      url={https://arxiv.org/abs/2411.12580}, 
}

@misc{han2020explainingblackboxpredictions,
      title={Explaining Black Box Predictions and Unveiling Data Artifacts through Influence Functions}, 
      author={Xiaochuang Han and Byron C. Wallace and Yulia Tsvetkov},
      year={2020},
      eprint={2005.06676},
      archivePrefix={arXiv},
      primaryClass={cs.CL},
      url={https://arxiv.org/abs/2005.06676}, 
}

@misc{koh2020ifs,
      title={Understanding Black-box Predictions via Influence Functions}, 
      author={Pang Wei Koh and Percy Liang},
      year={2020},
      eprint={1703.04730},
      archivePrefix={arXiv},
      primaryClass={stat.ML},
      url={https://arxiv.org/abs/1703.04730}, 
}

@misc{liu2025understandingr1zeroliketrainingcritical,
      title={Understanding R1-Zero-Like Training: A Critical Perspective}, 
      author={Zichen Liu and Changyu Chen and Wenjun Li and Penghui Qi and Tianyu Pang and Chao Du and Wee Sun Lee and Min Lin},
      year={2025},
      eprint={2503.20783},
      archivePrefix={arXiv},
      primaryClass={cs.LG},
      url={https://arxiv.org/abs/2503.20783}, 
}

@misc{engstrom2025optimizingmltrainingmetagradient,
      title={Optimizing ML Training with Metagradient Descent}, 
      author={Logan Engstrom and Andrew Ilyas and Benjamin Chen and Axel Feldmann and William Moses and Aleksander Madry},
      year={2025},
      eprint={2503.13751},
      archivePrefix={arXiv},
      primaryClass={stat.ML},
      url={https://arxiv.org/abs/2503.13751}, 
}

@inproceedings{modeldiff,
  title={Modeldiff: A framework for comparing learning algorithms},
  author={Shah, Harshay and Park, Sung Min and Ilyas, Andrew and Madry, Aleksander},
  booktitle={International Conference on Machine Learning},
  pages={30646--30688},
  year={2023},
  organization={PMLR}
}

@misc{jia2021scalabilityvsutilitysacrifice,
      title={Scalability vs. Utility: Do We Have to Sacrifice One for the Other in Data Importance Quantification?}, 
      author={Ruoxi Jia and Fan Wu and Xuehui Sun and Jiacen Xu and David Dao and Bhavya Kailkhura and Ce Zhang and Bo Li and Dawn Song},
      year={2021},
      eprint={1911.07128},
      archivePrefix={arXiv},
      primaryClass={cs.LG},
      url={https://arxiv.org/abs/1911.07128}, 
}

@misc{chang2024scalableinfluencefacttracing,
      title={Scalable Influence and Fact Tracing for Large Language Model Pretraining}, 
      author={Tyler A. Chang and Dheeraj Rajagopal and Tolga Bolukbasi and Lucas Dixon and Ian Tenney},
      year={2024},
      eprint={2410.17413},
      archivePrefix={arXiv},
      primaryClass={cs.CL},
      url={https://arxiv.org/abs/2410.17413}, 
}

@misc{akyurek2022tracingfactualknowledgelanguage,
      title={Towards Tracing Factual Knowledge in Language Models Back to the Training Data}, 
      author={Ekin Akyürek and Tolga Bolukbasi and Frederick Liu and Binbin Xiong and Ian Tenney and Jacob Andreas and Kelvin Guu},
      year={2022},
      eprint={2205.11482},
      archivePrefix={arXiv},
      primaryClass={cs.CL},
      url={https://arxiv.org/abs/2205.11482}, 
}

@inproceedings{zheng-jiang-2022-empirical,
    title = "An Empirical Study of Memorization in {NLP}",
    author = "Zheng, Xiaosen  and
      Jiang, Jing",
    editor = "Muresan, Smaranda  and
      Nakov, Preslav  and
      Villavicencio, Aline",
    booktitle = "Proceedings of the 60th Annual Meeting of the Association for Computational Linguistics (Volume 1: Long Papers)",
    month = may,
    year = "2022",
    address = "Dublin, Ireland",
    publisher = "Association for Computational Linguistics",
    url = "https://aclanthology.org/2022.acl-long.434/",
    doi = "10.18653/v1/2022.acl-long.434",
    pages = "6265--6278"
}

@misc{grosse2023studyinglargelanguagemodel,
      title={Studying Large Language Model Generalization with Influence Functions}, 
      author={Roger Grosse and Juhan Bae and Cem Anil and Nelson Elhage and Alex Tamkin and Amirhossein Tajdini and Benoit Steiner and Dustin Li and Esin Durmus and Ethan Perez and Evan Hubinger and Kamilė Lukošiūtė and Karina Nguyen and Nicholas Joseph and Sam McCandlish and Jared Kaplan and Samuel R. Bowman},
      year={2023},
      eprint={2308.03296},
      archivePrefix={arXiv},
      primaryClass={cs.LG},
      url={https://arxiv.org/abs/2308.03296}, 
}

@misc{lambert2025tulu3pushingfrontiers,
      title={Tulu 3: Pushing Frontiers in Open Language Model Post-Training}, 
      author={Nathan Lambert and Jacob Morrison and Valentina Pyatkin and Shengyi Huang and Hamish Ivison and Faeze Brahman and Lester James V. Miranda and Alisa Liu and Nouha Dziri and Shane Lyu and Yuling Gu and Saumya Malik and Victoria Graf and Jena D. Hwang and Jiangjiang Yang and Ronan Le Bras and Oyvind Tafjord and Chris Wilhelm and Luca Soldaini and Noah A. Smith and Yizhong Wang and Pradeep Dasigi and Hannaneh Hajishirzi},
      year={2025},
      eprint={2411.15124},
      archivePrefix={arXiv},
      primaryClass={cs.CL},
      url={https://arxiv.org/abs/2411.15124}, 
}

@misc{xia2024lessselectinginfluentialdata,
      title={LESS: Selecting Influential Data for Targeted Instruction Tuning}, 
      author={Mengzhou Xia and Sadhika Malladi and Suchin Gururangan and Sanjeev Arora and Danqi Chen},
      year={2024},
      eprint={2402.04333},
      archivePrefix={arXiv},
      primaryClass={cs.CL},
      url={https://arxiv.org/abs/2402.04333}, 
}

@inproceedings{trak,
    author = {Park, Sung Min and Georgiev, Kristian and Ilyas, Andrew and Leclerc, Guillaume and M\k{a}dry, Aleksander},
    title = {TRAK: attributing model behavior at scale},
    year = {2023},
    publisher = {JMLR.org},
    booktitle = {Proceedings of the 40th International Conference on Machine Learning},
    articleno = {1128},
    numpages = {40},
    location = {Honolulu, Hawaii, USA},
    series = {ICML'23}
}

@misc{yeh2018representerpointselectionexplaining,
      title={Representer Point Selection for Explaining Deep Neural Networks}, 
      author={Chih-Kuan Yeh and Joon Sik Kim and Ian E. H. Yen and Pradeep Ravikumar},
      year={2018},
      eprint={1811.09720},
      archivePrefix={arXiv},
      primaryClass={cs.LG},
      url={https://arxiv.org/abs/1811.09720}, 
}

@misc{charpiat2021inputsimilarityneuralnetwork,
      title={Input Similarity from the Neural Network Perspective}, 
      author={Guillaume Charpiat and Nicolas Girard and Loris Felardos and Yuliya Tarabalka},
      year={2021},
      eprint={2102.05262},
      archivePrefix={arXiv},
      primaryClass={cs.LG},
      url={https://arxiv.org/abs/2102.05262}, 
}

@misc{ilyas2025magic,
      title={MAGIC: Near-Optimal Data Attribution for Deep Learning}, 
      author={Andrew Ilyas and Logan Engstrom},
      year={2025},
      eprint={2504.16430},
      archivePrefix={arXiv},
      primaryClass={cs.LG},
      url={https://arxiv.org/abs/2504.16430}, 
}

@inproceedings{wolf2020transformers,
  title={Transformers: State-of-the-art natural language processing},
  author={Wolf, Thomas and Debut, Lysandre and Sanh, Victor and Chaumond, Julien and Delangue, Clement and Moi, Anthony and Cistac, Pierric and Rault, Tim and Louf, R{\'e}mi and Funtowicz, Morgan and others},
  booktitle={Proceedings of the 2020 conference on empirical methods in natural language processing: system demonstrations},
  pages={38--45},
  year={2020}
}

@incollection{shapley1953value,
  title = {A Value for n-Person Games},
  author = {Shapley, Lloyd S},
  booktitle = {Contributions to the Theory of Games II},
  editor = {Kuhn, Harold W. and Tucker, Albert W.},
  pages = {307--317},
  year = {1953},
  publisher = {Princeton University Press},
  address = {Princeton}
}

@misc{bae2024sourceunrolled,
      title={Training Data Attribution via Approximate Unrolled Differentiation}, 
      author={Juhan Bae and Wu Lin and Jonathan Lorraine and Roger Grosse},
      year={2024},
      eprint={2405.12186},
      archivePrefix={arXiv},
      primaryClass={cs.LG},
      url={https://arxiv.org/abs/2405.12186}, 
}

@article{shumailov2021manipulating,
  title={Manipulating sgd with data ordering attacks},
  author={Shumailov, Ilia and Shumaylov, Zakhar and Kazhdan, Dmitry and Zhao, Yiren and Papernot, Nicolas and Erdogdu, Murat A and Anderson, Ross J},
  journal={Advances in Neural Information Processing Systems},
  volume={34},
  pages={18021--18032},
  year={2021}
}

@inproceedings{jaburi2025mitigating,
  title={Mitigating Emergent Misalignment with Data Attribution},
  author={Jaburi, Louis and Paulo, Gon{\c{c}}alo and Shabalin, Stepan and Quirke, Lucia and Belrose, Nora},
  booktitle={Mechanistic Interpretability Workshop at NeurIPS 2025},
  year={2025}
}

@misc{zheng2023judgingllms,
      title={Judging LLM-as-a-Judge with MT-Bench and Chatbot Arena}, 
      author={Lianmin Zheng and Wei-Lin Chiang and Ying Sheng and Siyuan Zhuang and Zhanghao Wu and Yonghao Zhuang and Zi Lin and Zhuohan Li and Dacheng Li and Eric P. Xing and Hao Zhang and Joseph E. Gonzalez and Ion Stoica},
      year={2023},
      eprint={2306.05685},
      archivePrefix={arXiv},
      primaryClass={cs.CL},
      url={https://arxiv.org/abs/2306.05685}, 
}

@misc{wei2022emergentabilities,
      title={Emergent Abilities of Large Language Models}, 
      author={Jason Wei and Yi Tay and Rishi Bommasani and Colin Raffel and Barret Zoph and Sebastian Borgeaud and Dani Yogatama and Maarten Bosma and Denny Zhou and Donald Metzler and Ed H. Chi and Tatsunori Hashimoto and Oriol Vinyals and Percy Liang and Jeff Dean and William Fedus},
      year={2022},
      eprint={2206.07682},
      archivePrefix={arXiv},
      primaryClass={cs.CL},
      url={https://arxiv.org/abs/2206.07682}, 
}

@misc{zhang2024simplefsdp,
      title={SimpleFSDP: Simpler Fully Sharded Data Parallel with torch.compile}, 
      author={Ruisi Zhang and Tianyu Liu and Will Feng and Andrew Gu and Sanket Purandare and Wanchao Liang and Francisco Massa},
      year={2024},
      eprint={2411.00284},
      archivePrefix={arXiv},
      primaryClass={cs.DC},
      url={https://arxiv.org/abs/2411.00284}, 
}

@misc{mlodozeniec2025dtda,
      title={Distributional Training Data Attribution: What do Influence Functions Sample?}, 
      author={Bruno Mlodozeniec and Isaac Reid and Sam Power and David Krueger and Murat Erdogdu and Richard E. Turner and Roger Grosse},
      year={2025},
      eprint={2506.12965},
      archivePrefix={arXiv},
      primaryClass={cs.LG},
      url={https://arxiv.org/abs/2506.12965}, 
}

@misc{biderman2023pythiasuite,
      title={Pythia: A Suite for Analyzing Large Language Models Across Training and Scaling}, 
      author={Stella Biderman and Hailey Schoelkopf and Quentin Anthony and Herbie Bradley and Kyle O'Brien and Eric Hallahan and Mohammad Aflah Khan and Shivanshu Purohit and USVSN Sai Prashanth and Edward Raff and Aviya Skowron and Lintang Sutawika and Oskar van der Wal},
      year={2023},
      eprint={2304.01373},
      archivePrefix={arXiv},
      primaryClass={cs.CL},
      url={https://arxiv.org/abs/2304.01373}, 
}

@misc{touvron2023llama,
      title={LLaMA: Open and Efficient Foundation Language Models}, 
      author={Hugo Touvron and Thibaut Lavril and Gautier Izacard and Xavier Martinet and Marie-Anne Lachaux and Timothée Lacroix and Baptiste Rozière and Naman Goyal and Eric Hambro and Faisal Azhar and Aurelien Rodriguez and Armand Joulin and Edouard Grave and Guillaume Lample},
      year={2023},
      eprint={2302.13971},
      archivePrefix={arXiv},
      primaryClass={cs.CL},
      url={https://arxiv.org/abs/2302.13971}, 
}

@misc{yang2025qwen3,
      title={Qwen3 Technical Report}, 
      author={An Yang and Anfeng Li and Baosong Yang and Beichen Zhang and Binyuan Hui and Bo Zheng and Bowen Yu and Chang Gao and Chengen Huang and Chenxu Lv and Chujie Zheng and Dayiheng Liu and Fan Zhou and Fei Huang and Feng Hu and Hao Ge and Haoran Wei and Huan Lin and Jialong Tang and Jian Yang and Jianhong Tu and Jianwei Zhang and Jianxin Yang and Jiaxi Yang and Jing Zhou and Jingren Zhou and Junyang Lin and Kai Dang and Keqin Bao and Kexin Yang and Le Yu and Lianghao Deng and Mei Li and Mingfeng Xue and Mingze Li and Pei Zhang and Peng Wang and Qin Zhu and Rui Men and Ruize Gao and Shixuan Liu and Shuang Luo and Tianhao Li and Tianyi Tang and Wenbiao Yin and Xingzhang Ren and Xinyu Wang and Xinyu Zhang and Xuancheng Ren and Yang Fan and Yang Su and Yichang Zhang and Yinger Zhang and Yu Wan and Yuqiong Liu and Zekun Wang and Zeyu Cui and Zhenru Zhang and Zhipeng Zhou and Zihan Qiu},
      year={2025},
      eprint={2505.09388},
      archivePrefix={arXiv},
      primaryClass={cs.CL},
      url={https://arxiv.org/abs/2505.09388}, 
}

@misc{obrien2026deepignorance,
      title={Deep Ignorance: Filtering Pretraining Data Builds Tamper-Resistant Safeguards into Open-Weight LLMs}, 
      author={Kyle O'Brien and Stephen Casper and Quentin Anthony and Tomek Korbak and Robert Kirk and Xander Davies and Ishan Mishra and Geoffrey Irving and Yarin Gal and Stella Biderman},
      year={2026},
      eprint={2508.06601},
      archivePrefix={arXiv},
      primaryClass={cs.LG},
      url={https://arxiv.org/abs/2508.06601}, 
}

@misc{li2024wmdp,
      title={The WMDP Benchmark: Measuring and Reducing Malicious Use With Unlearning}, 
      author={Nathaniel Li and Alexander Pan and Anjali Gopal and Summer Yue and Daniel Berrios and Alice Gatti and Justin D. Li and Ann-Kathrin Dombrowski and Shashwat Goel and Long Phan and Gabriel Mukobi and Nathan Helm-Burger and Rassin Lababidi and Lennart Justen and Andrew B. Liu and Michael Chen and Isabelle Barrass and Oliver Zhang and Xiaoyuan Zhu and Rishub Tamirisa and Bhrugu Bharathi and Adam Khoja and Zhenqi Zhao and Ariel Herbert-Voss and Cort B. Breuer and Samuel Marks and Oam Patel and Andy Zou and Mantas Mazeika and Zifan Wang and Palash Oswal and Weiran Lin and Adam A. Hunt and Justin Tienken-Harder and Kevin Y. Shih and Kemper Talley and John Guan and Russell Kaplan and Ian Steneker and David Campbell and Brad Jokubaitis and Alex Levinson and Jean Wang and William Qian and Kallol Krishna Karmakar and Steven Basart and Stephen Fitz and Mindy Levine and Ponnurangam Kumaraguru and Uday Tupakula and Vijay Varadharajan and Ruoyu Wang and Yan Shoshitaishvili and Jimmy Ba and Kevin M. Esvelt and Alexandr Wang and Dan Hendrycks},
      year={2024},
      eprint={2403.03218},
      archivePrefix={arXiv},
      primaryClass={cs.LG},
      url={https://arxiv.org/abs/2403.03218}, 
}

@misc{pruthi2020estimatingtrainingdatainfluence,
      title={Estimating Training Data Influence by Tracing Gradient Descent}, 
      author={Garima Pruthi and Frederick Liu and Mukund Sundararajan and Satyen Kale},
      year={2020},
      eprint={2002.08484},
      archivePrefix={arXiv},
      primaryClass={cs.LG},
      url={https://arxiv.org/abs/2002.08484}, 
}

@misc{kwon2024datainfefficientlyestimatingdata,
      title={DataInf: Efficiently Estimating Data Influence in LoRA-tuned LLMs and Diffusion Models}, 
      author={Yongchan Kwon and Eric Wu and Kevin Wu and James Zou},
      year={2024},
      eprint={2310.00902},
      archivePrefix={arXiv},
      primaryClass={cs.LG},
      url={https://arxiv.org/abs/2310.00902}, 
}

@inproceedings{Martens2010DeepLV,
  title={Deep learning via Hessian-free optimization},
  author={James Martens},
  booktitle={International Conference on Machine Learning},
  year={2010},
  url={https://api.semanticscholar.org/CorpusID:11154521}
}

@misc{schioppa2021scalinginfluencefunctions,
      title={Scaling Up Influence Functions}, 
      author={Andrea Schioppa and Polina Zablotskaia and David Vilar and Artem Sokolov},
      year={2021},
      eprint={2112.03052},
      archivePrefix={arXiv},
      primaryClass={cs.LG},
      url={https://arxiv.org/abs/2112.03052}, 
}

@misc{agarwal2017secondorderstochasticoptimizationmachine,
      title={Second-Order Stochastic Optimization for Machine Learning in Linear Time}, 
      author={Naman Agarwal and Brian Bullins and Elad Hazan},
      year={2017},
      eprint={1602.03943},
      archivePrefix={arXiv},
      primaryClass={stat.ML},
      url={https://arxiv.org/abs/1602.03943}, 
}

@misc{hu2021loralowrankadaptationlarge,
      title={LoRA: Low-Rank Adaptation of Large Language Models}, 
      author={Edward J. Hu and Yelong Shen and Phillip Wallis and Zeyuan Allen-Zhu and Yuanzhi Li and Shean Wang and Lu Wang and Weizhu Chen},
      year={2021},
      eprint={2106.09685},
      archivePrefix={arXiv},
      primaryClass={cs.CL},
      url={https://arxiv.org/abs/2106.09685}, 
}

@software{biderman2024lmeval,
       author = {{Gao}, Leo and {Tow}, Jonathan and {Biderman}, Stella and {Black}, Sid and {DiPofi}, Anthony and {Foster}, Charles and {Golding}, Laurence and {Hsu}, Jeffrey and {McDonell}, Kyle and {Muennighoff}, Niklas and {Phang}, Jason and {Reynolds}, Laria and {Tang}, Eric and {Thite}, Anish and {Wang}, Ben and {Wang}, Kevin and {Zou}, Andy},
        title = "{A framework for few-shot language model evaluation}",
         year = 2021,
        month = sep,
          eid = {10.5281/zenodo.5371629},
          doi = {10.5281/zenodo.5371629},
      version = {v0.0.1},
    publisher = {Zenodo},
       adsurl = {https://ui.adsabs.harvard.edu/abs/2021zndo...5371629G}
}

@misc{shao2024deepseekmath,
      title={DeepSeekMath: Pushing the Limits of Mathematical Reasoning in Open Language Models}, 
      author={Zhihong Shao and Peiyi Wang and Qihao Zhu and Runxin Xu and Junxiao Song and Xiao Bi and Haowei Zhang and Mingchuan Zhang and Y. K. Li and Y. Wu and Daya Guo},
      year={2024},
      eprint={2402.03300},
      archivePrefix={arXiv},
      primaryClass={cs.CL},
      url={https://arxiv.org/abs/2402.03300}, 
}

@inproceedings{lesci2024causal,
  title={Causal estimation of memorisation profiles},
  author={Lesci, Pietro and Meister, Clara and Hofmann, Thomas and Vlachos, Andreas and Pimentel, Tiago},
  booktitle={Proceedings of the 62nd Annual Meeting of the Association for Computational Linguistics (Volume 1: Long Papers)},
  pages={15616--15635},
  year={2024}
}

@Misc{peft,
  title =        {{PEFT}: State-of-the-art Parameter-Efficient Fine-Tuning methods},
  author =       {Sourab Mangrulkar and Sylvain Gugger and Lysandre Debut and Younes Belkada and Sayak Paul and Benjamin Bossan and Marian Tietz},
  howpublished = {\url{https://github.com/huggingface/peft}},
  year =         {2022}
}

@misc{brown2020llmsfewshotlearners,
      title={Language Models are Few-Shot Learners}, 
      author={Tom B. Brown and Benjamin Mann and Nick Ryder and Melanie Subbiah and Jared Kaplan and Prafulla Dhariwal and Arvind Neelakantan and Pranav Shyam and Girish Sastry and Amanda Askell and Sandhini Agarwal and Ariel Herbert-Voss and Gretchen Krueger and Tom Henighan and Rewon Child and Aditya Ramesh and Daniel M. Ziegler and Jeffrey Wu and Clemens Winter and Christopher Hesse and Mark Chen and Eric Sigler and Mateusz Litwin and Scott Gray and Benjamin Chess and Jack Clark and Christopher Berner and Sam McCandlish and Alec Radford and Ilya Sutskever and Dario Amodei},
      year={2020},
      eprint={2005.14165},
      archivePrefix={arXiv},
      primaryClass={cs.CL},
      url={https://arxiv.org/abs/2005.14165}, 
}

@misc{rathi2026tokenleveldata,
      title={Shaping capabilities with token-level data filtering}, 
      author={Neil Rathi and Alec Radford},
      year={2026},
      eprint={2601.21571},
      archivePrefix={arXiv},
      primaryClass={cs.LG},
      url={https://arxiv.org/abs/2601.21571}, 
}

@misc{wan2025tokenneedsforgetting,
      title={Not Every Token Needs Forgetting: Selective Unlearning to Limit Change in Utility in Large Language Model Unlearning}, 
      author={Yixin Wan and Anil Ramakrishna and Kai-Wei Chang and Volkan Cevher and Rahul Gupta},
      year={2025},
      eprint={2506.00876},
      archivePrefix={arXiv},
      primaryClass={cs.CL},
      url={https://arxiv.org/abs/2506.00876}, 
}

@misc{han2024wildguard,
      title={WildGuard: Open One-Stop Moderation Tools for Safety Risks, Jailbreaks, and Refusals of LLMs}, 
      author={Seungju Han and Kavel Rao and Allyson Ettinger and Liwei Jiang and Bill Yuchen Lin and Nathan Lambert and Yejin Choi and Nouha Dziri},
      year={2024},
      eprint={2406.18495},
      archivePrefix={arXiv},
      primaryClass={cs.CL},
      url={https://arxiv.org/abs/2406.18495}, 
}

% \newpage
\clearpage

\appendix

\section{Optimizations}
\label{sec:appendix}

\paragraph{MAGIC checkpoint strategies}\label{magic_checkpoint}

Bergson implements the \textsc{MAGIC} algorithm, which uses $O(\log N)$ disk space, $O(\log N)$ memory, and $O(N \log N)$ computation, and a novel alternative algorithm which uses only $O(N)$ computation at the cost of $O(\sqrt N)$ memory and $O(\sqrt N)$ disk space.

\paragraph{TrackStar gradient batching}

We group documents into minibatches to speed up the computation of gradients. For efficiency, we attempt to minimize the number of padding tokens used by grouping together documents of similar lengths, similar to the \texttt{group\_by\_length} option on the HuggingFace transformers \texttt{Trainer} class \citep{wolf2020transformers}. Our allocation algorithm assigns different numbers of sequences to different batches, while ensuring that the total number of tokens in each batch is roughly constant, and never exceeds a specified threshold.

\paragraph{(E)K-FAC memory management}\label{ekfac_memory}

Uncompressed Hessian approximations such as (E)K-FAC require many times more memory than the model parameters. We ameliorate these costs by sharding the approximations across devices, enabling larger batches and lower overall computational costs. 
%(\louis{todo: should measure this, how?}

\section{Feature Support}

\paragraph{Training}
Bergson provides a differentiable, deterministic, and scalable model trainer that enables both unrolled differentiation and fixed-seed attribution. Sequences or tokens can be filtered from training without changing data order or batch composition by re-weighting their losses to zero using a weighted loss function. This results in a training setup that more closely matches the assumptions of the influence function derivation.

\paragraph{Gradients}

We support gradient collection from linear (weight and bias) parameters using KL, CE, and GRPO losses. We also support gradient compression using random projection matrices sampled from Rademacher or uniform distributions, and single- or double-sided projection. A HuggingFace Trainer Callback is provided for the collection of raw training gradients, at an estimated runtime overhead of 17\%.

\paragraph{Reproducibility}

Each tool in Bergson is configured using dataclasses populated with sensible defaults, and each command serializes its configuration dataclasses when called. The resulting artifact may be used with the CLI to launch a replication. We provide command-line and programmatic access to each tool.

% MAGIC differentiates through the same operations used in model training, so its distributed requirements are similar---we implement a differentiable Trainer with DDP and FSDP support using SimpleFSDP 

\paragraph{Scaling}\label{distributed_ops}
Bergson natively supports DDP and FSDP for multi-node attribution, with configuration options modeled after \texttt{torchrun}. We use SimpleFSDP to unlock the MAGIC double backwards pass \citep{zhang2024simplefsdp}. Because influence function computations do not require communication between FSDP groups, very large-scale multi-node runs where each FSDP group independently processes a dataset shard are straightforward, and a demonstration is provided.

% \will{make this more user facing, ie what they can do, and what they don't have to do, otherwise move to appx and add a hook in supported workflow}

% This really depends on what you want to do
% If you want to do MAGIC you basically get what you get
% If you want to do K-FAC/influence functions you need to compute hessians upfront, and we provide parallelism to make this easy
% If you want to build an index or run prepared influence functions it's just like training

% Memory usage in influence functions differs from model training in three ways. First, memory requirements are reduced when collecting compressed gradients because the full parameter gradients are not instantiated, enabling larger-scale operations. Second, optimizer states such as Adam buffers may be omitted for many methods. Finally, influence functions such as EK-FAC prepare Hessian approximations, which require significant memory. 

% We find that block-diagonal Hessian approximations, such as those in the K-FAC family, may be profitably sharded block-wise across devices during estimation, similar to the sharding of model layers in fully sharded data parallel (FSDP) model training (see \ref{ekfac_memory}.) 

\paragraph{Usability}
Bergson integrates with HuggingFace Transformers and Datasets, and can load on-disk datasets in a variety of formats. To support attribution of evaluation sets with diverse formatting and precise label masking requirements, we follow the LM Evaluation Harness in providing a YAML templating system and a standard template for MCQA-style evaluations \citep{biderman2024lmeval}. We provide utilities including an automatic batch sizing tool and a healthcheck tool that ensures attribution setups are numerically stable and suggests fixes for common issues. Finally, we provide several examples demonstrating library feature applications, such as data poisoning detection, LESS data curation, and attributing induction.

\section{Library Comparison}

We compare Bergson to two major alternative libraries for data attribution: Kronfluence and dattri. We compare libraries in terms of supported methods (Table \ref{tab:method_families}) and other features (Table \ref{tab:feature_support}).
% ---------------------------------------------------------------------
%  feature support (full width)
% ---------------------------------------------------------------------
\begin{table*}[t]
\centering
\small
\setlength{\tabcolsep}{5pt}
\begin{tabular}{lccc}
\toprule
\textbf{Capability} & \textbf{Bergson} & \textbf{Kronflu.} & \textbf{dattri} \\
\midrule
Optimizer norm.\ (Adam/Adafactor)   & \yes                    & \no  & \no  \\
On-disk gradient store              & \yes                    & \no  & \prt\textsuperscript{a} \\
ANN index (FAISS)                   & \yes                    & \no  & \no  \\
Gradient compression                & \yes\textsuperscript{b} & \yes\textsuperscript{c} & \yes\textsuperscript{b} \\
Per-token attribution               & \yes                    & \yes & \no  \\
Multi-GPU (FSDP)                     & \yes                    & \yes\textsuperscript{d} & \no  \\
Multi-node (FSDP)                    & \yes                    & \yes & \no  \\
RL objectives (GRPO)                 & \yes                    & \no  & \no  \\
Packaged benchmark suite            & \no                     & \no  & \yes \\
Built-in LDS evaluation             & \yes                    & \no & \yes \\
MoE expert attribution\textsuperscript{f} & \no               & \no  & \yes \\
\midrule
\multicolumn{4}{l}{\textbf{Largest single-node model by method:}}\\
grad-dot (identity)            & \textbf{72B}            & 32B  & 1.5B \\
EK-FAC (Kronecker)             & \textbf{14B} & \textbf{14B}  & 0.77B\textsuperscript{g} \\
\bottomrule
\end{tabular}
\caption{Validated feature support across data attribution libraries. The scale rows indicate the largest model whose full attribution completed off-the-shelf on one 8$\times$A100 node, which is strongly method dependent. \yes\ verified; \prt\ partial;
\no\ unsupported.
\newline\textsuperscript{a}Crashed at batch size 5000, works at batch size 500.
\newline\textsuperscript{b}Random projection.
\newline\textsuperscript{c}Low-rank SVD.
\newline\textsuperscript{d}Kronfluence's \texttt{apply\_fsdp} helper fails on models with frozen parameters so we use \texttt{accelerate} FSDP.
\newline\textsuperscript{f}Bergson and Kronfluence do not track the experts or router in recent fused-parameter MoE layouts, but work for older per-expert \texttt{nn.Linear} layouts.
\newline\textsuperscript{g}dattri EK-FAC requires a patch to run on transformer (LM) inputs.}
\label{tab:feature_support}
\end{table*}

% ---------------------------------------------------------------------
%  TABLE 2 — method-family map (full width; appendix-friendly)
% ---------------------------------------------------------------------
\begin{table*}[t]
\centering
\small
\begin{tabular}{llccc}
\toprule
\textbf{Family} & \textbf{Method} & \textbf{Bergson} & \textbf{Kronflu.} & \textbf{dattri} \\
\midrule
\multicolumn{5}{l}{\emph{Unrolled differentiation}}\\
& Exact (MAGIC)                       & \yes & \no & \no \\
& Approximate unrolling (SOURCE) & \yes & \no & \no \\
& Approximate unrolling (Data Value Embedding) & \no & \no & \yes \\
\addlinespace
\multicolumn{5}{l}{\emph{Influence functions --- by Hessian approximation}}\\
& grad-dot (identity)                 & \yes & \yes & \yes \\
& grad-cosine (identity + unit-norm)  & \yes & \no  & \yes \\
& K-FAC / EK-FAC         & \yes & \yes & \yes \\
& Flat (full) autocorrelation         & \yes & \no  & \yes \\
& Shampoo / TK-FAC                    & \yes & \no  & \no \\
& Iterative solvers (LiSSA/CG/Arnoldi)& \no  & \no  & \yes \\
& DataInf (closed-form)               & \no  & \no  & \yes \\
& RelatIF                             & \no  & \no  & \yes \\
& LoGra (low-rank proj.) & \no  & \no  & \yes \\
\addlinespace
\multicolumn{5}{l}{\emph{Other}}\\
& TracIn (multi-checkpoint)           & \no  & \no  & \yes \\
& TRAK                                & \yes  & \no & \yes \\
& RPS (representer points)            & \no  & \no  & \yes \\
& KNN-Shapley / valuation             & \no  & \no  & \yes \\
& GraSS (sparse proj.) & \no  & \no  & \yes\textsuperscript{b} \\
\bottomrule
\end{tabular}
\caption{Method families and library coverage.
\newline\textsuperscript{b}We did not perform a GraSS test run because it used a kernel incompatible with our version of CUDA.}
\label{tab:method_families}
\end{table*}

\end{document}